\documentclass{article}

\usepackage{amsmath,amsfonts}
\usepackage{algorithmic}
\usepackage{array}
\usepackage[caption=false,font=normalsize,labelfont=sf,textfont=sf]{subfig}
\usepackage{amssymb}
\usepackage{textcomp}
\usepackage{stfloats}
\usepackage{url}
\usepackage{verbatim}
\usepackage{graphicx}
\usepackage{hyperref}
\usepackage{siunitx}
\usepackage{pdflscape}
\usepackage{longtable}
\usepackage{booktabs}
\usepackage{colortbl}
\usepackage{caption}
\usepackage[table]{xcolor}
\usepackage{xspace}
\usepackage{makecell}
\usepackage{array}
\usepackage{scalerel}
\usepackage{tikz}
\usepackage{multirow}
\usepackage{multicol}
\usepackage{ifthen}
\usepackage{wrapfig}

\usepackage{collcell}
\makeatletter
\newcolumntype{G}{>{\collectcell\@gobble}c<{\endcollectcell}@{}}
\makeatother

\usepackage{booktabs}

\newcommand{\argmin}{\operatornamewithlimits{argmin}}

\newcommand{\result}[4]{${#1}_{\scaleto{\pm #2}{3pt}}{\scaleto{\vert \raisebox{.8pt}{-}#3_{\scaleto{\pm #4}{3pt}}}{8pt}}$}

\newcommand{\resultp}[4]{${#1}_{\scaleto{\pm #2}{3pt}}{\scaleto{\vert \raisebox{.8pt}{+}#3_{\scaleto{\pm #4}{3pt}}}{8pt}}$}

\newcommand\mybox[2][]{\tikz[overlay]\node[fill=blue!20,inner sep=2pt, anchor=text, rectangle, rounded corners=1mm,#1] {#2};\phantom{#2}}

\newcommand{\nhead}{Cascaded Gates\xspace}
\newcommand{\nheada}{CG\xspace}

\newcommand{\nreg}{Margin Dampening\xspace}
\newcommand{\nrega}{MD\xspace}

\newcommand{\napproach}{\nrega-\nheada}

\def\BibTeX{{\rm B\kern-.05em{\sc i\kern-.025em b}\kern-.08em
    T\kern-.1667em\lower.7ex\hbox{E}\kern-.125emX}}
\usepackage[nonatbib, preprint]{neurips_2024}
\usepackage[numbers]{natbib}

\title{Class incremental learning with probability dampening and cascaded gated classifier}

\author{%
 Jary Pomponi\thanks{Corresponding author} \quad
  Alessio Devoto \quad
   Simone Scardapane \\ 
 Information Engineering, Electronics and Telecommunications (DIET) \\
  Sapienza University of Rome, Italy \\
  \texttt{\{name\}.\{surname\}@uniroma1.it}
}
\begin{document}

\maketitle

\begin{abstract}
Humans are capable of acquiring new knowledge and transferring learned knowledge into different domains, incurring a small forgetting. The same ability, called Continual Learning, is challenging to achieve when operating with neural networks due to the forgetting affecting past learned tasks when learning new ones. This forgetting can be mitigated by replaying stored samples from past tasks, but a large memory size may be needed for long sequences of tasks; moreover, this could lead to overfitting on saved samples. In this paper, we propose a novel regularisation approach and a novel incremental classifier called, respectively, \nreg and \nhead. The first combines a constraining loss 
and a knowledge distillation approach to preserve past learned knowledge while allowing the model to learn new patterns effectively. The latter is a gated incremental classifier, helping the model modify past predictions without directly interfering with them. This is achieved by modifying the output of the model with auxiliary scaling functions.
We empirically show that our approach performs well on multiple benchmarks against well-established baselines, and we also study each component of our proposal and how the combinations of such components affect the final results.

\end{abstract}

\section{Introduction}
%
\begin{wrapfigure}{TR}{0.45\textwidth}
  \centering
\includegraphics[width=1\linewidth]{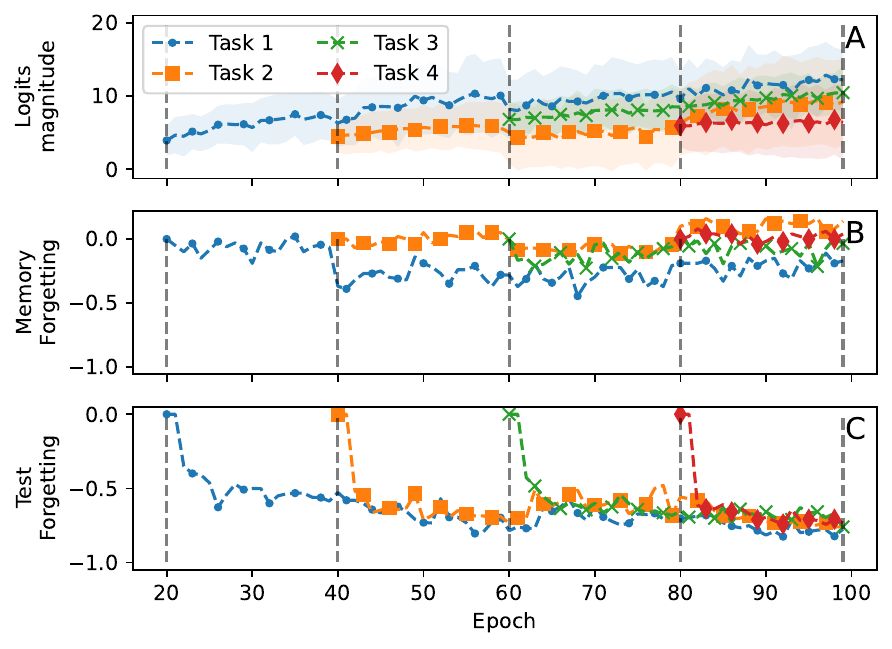}
    \caption{We train a ResNet-20 model with a simple rehearsal approach and a memory size of 500, on CIFAR10 divided into 5 tasks. A) Magnitude of ground truth logits on the replay samples; B) Forgetting on the replay samples; C) Forgetting on test scores for each past task.}
    \label{fig:bias_intro}
\end{wrapfigure}
%
Building an agent capable of continuously learning over time from a stream of tasks remains a fundamental challenge. This ability is called Continual Learning (CL), a key property enabling autonomous real-world agents.
Traditional machine learning models often struggle when faced with sequential and non-stationary data due to Catastrophic Forgetting (CF), which often leads to a drastic performance decrease due to the overwriting of past learned knowledge while learning new one \cite{PARISI201954}. The desired balance between remembering past learned patterns and correctly acquiring new knowledge is often called \emph{stability-plasticity} trade-off \cite{wang2023comprehensive}. 

The most straightforward way to mitigate CF is by using rehearsal-based approaches, which store and replay a small subset of previous task samples, but how such samples must be saved and retrieved is an open question. It is challenging to keep generally discriminative past samples because of overfitting that arises from replaying them \cite{Verwimp_2021, prabhu2020gdumb, buzzega2020dark, bonicelli2022effectiveness}. In such scenarios, additional complexity comes from the intrinsically unbalanced nature of the training procedure, in which newer patterns have more samples than the ones in the memory.
Consider the example shown in Fig. \ref{fig:bias_intro}. 
The ground truth logits produced by the model grow during the whole training procedure, exposing an overfitting phenomena on rehearsal samples. 
At the same time, the forgetting on the test samples is higher than the one achieved on rehearsal samples. Intuitively, these two quantities should be similar, showing that this approach is not fighting the CF but only overfitting patterns present in the external memory. This is evident by looking at scores achieved for tasks 3 and 4, in which forgetting on replay samples is nearly zero across all the training, and often becomes positive, while the accuracy on the corresponding test sets rapidly decreases. Despite such limitations, using past samples as training ones is one of the most used approaches to fight the CF. 

Usually, a rehearsal approach is coupled with a regularisation strategy, in which past samples are used to calculate a regularisation term added to the training loss to alleviate  CF further. In such scenarios, the true nature of the forgetting still needs to be understood. Some works claim it could arise from the classifier \cite{ahn2021ss, wu2019large, zhao2020maintaining, li2017learning}, and most of the time the output of the model is directly regularised to mitigate this. The regularisation is usually carried out by forcing the classifier to produce the same output distribution as in the past only for rehearsal samples. 
This combination creates two losses that are in contrast, because training on past samples leads to overfitting while regularising on them forces the model to produce the same output as in the past. Such approaches usually require a careful selection of the hyperparameters to balance these two aspects. This behaviour is shown in Appendix \ref{ap:reg}. 

Motivated by these observations, we propose a novel CL approach composed of a regularisation term and a novel incremental classifier head. The former modifies the probabilities of previous classes up to a certain margin which satisfies a given constraint while, at the same time, regularising all the classes using a Knowledge Distillation (KD) approach calculated over the whole output of the model. The latter is a hybrid classifier composed of smaller task-wise classifiers, which are scaled and combined to produce the final output distribution. The combination of such components is capable of correctly regularising past learned knowledge without training on past samples, avoiding overfitting of such samples. At the same time, current patterns are correctly learned without interfering with past learned ones. To demonstrate the effectiveness of our proposals, we conduct comprehensive experiments on multiple CL benchmarks, as well as in-depth exploratory experiments to analyse the effectiveness of core design choices of both the regularisation approach, called \nreg (\nrega), as well as the classification head, which we call \nhead (\nheada). For space reasons, an analysis of related works can be found in Appendix \ref{ap:rw}.

\section{Continual Learning definition}
\label{sec:cl_definition}

In a Class Incremental Learning (CIL) scenario, a generic model $f(\cdot)$ is trained on a sequence of classification tasks $\mathbf{T} = \{ \mathcal{T}_i \}$, with $i = 1, \dots, T$, having complete access only to the current training task $t$, with the additional possibility of storing a subset of past tasks' samples. In this paper we deal with image classification datasets, in which a generic task consists of a tuple $\mathcal{T}_i = (\mathcal{D}^i, \mathrm{Y}^t)$, where $\mathcal{D}^i = \{ (x_i, y_i)\}$ is the dataset containing tuples, each one composed by the sample $x$ and the scalar label
 $y_i \in \mathrm{Y}^t$, where $\mathrm{Y}^t$ is the label space of the task $t$, such that $\bigcap_{t = 1, \dots, \text{T}}\mathrm{Y}^t = \varnothing$; no other assumptions are made on the labels' space except that $\lvert \mathrm{Y}^t \rvert \ge 2$ for each task $t$.\footnote{We also assume that the classes are in a sequential order starting from zero, but this is just a scaling factor that has no impact on the scenario itself.} Moreover, by definition, in a CIL scenario task identities are present only during training but not in the inference phase. This key difference makes CIL harder than another CL scenario called Task Incremental Learning (TIL), because having access to the task identities during inference allows for better separation between tasks, drastically reducing CF effects. 
For this reason, TIL can be easily solved using an architectural approach \cite{golkar2019continual, wortsman2020supermasks} or even employing a regularisation approach without additional external memory \cite{pomponi2022centroids}. 

Formally, we aim to fit a model $f : \mathcal{X} \rightarrow \mathcal{Y}^{\text{T}}$, where $\mathcal{Y}^{\text{T}} = \bigcup_{t = 1, \dots, \text{T}}\mathrm{Y}^t $, which minimises the expected risk: 
\begin{equation} 
\label{eq:risk}
f^* = \argmin\limits_{f} \underset{(x, y) \sim \bigcup_{i = 1}^{\text{T}} \mathcal{D}^i}{\mathbb{E}} \, \mathcal{L}(f(x), y)
\end{equation}
\noindent where $\mathcal{L}$ stands for the loss (e.g. cross-entropy loss) of correctly classifying sample $x$ having label $y$ using the model $f$. A generic model $f$ used in a CL scenario can be further decomposed into a backbone $b(\cdot)$ and a classification head $h^t(\cdot)$, which can be conditioned to the task $t$,\footnote{If the classifier cannot be conditioned the task is ignored.} such that $f^t(x) = h^t(b(x))$.
No assumptions are made about the model, but usually, in a CIL scenario the backbone is given and its structure is fixed, while the output of the model is adapted each time a new task is retrieved. This is done by modifying the output of the classifier such that its output size is expanded to match the total number of classes, including the new ones;
we also operate in this setting. 

The goal of the minimisation process is to find the optimal model capable of minimising the loss over all the tasks. However, when training on a given task $t$ we cannot access the whole past tasks $t'<t$, making this minimisation infeasible. To overcome this constraint while training on a task $t>1$ we can use a regularisation function $\mathcal{R}$, an external memory $\mathcal{M}$ containing a small portion of past samples, or combine these two approaches, as such:
\begin{equation} 
\label{eq:risk_reg}
f^* = \argmin\limits_{f}\underset{(x, y) \sim \mathcal{D}^t \cup \mathcal{M}}{\mathbb{E}} \, \mathcal{L}(f(x), y) + \mathcal{R}(x, y)
\end{equation}
A well-crafted CL approach term prevents the model from forgetting past learned knowledge while leaving room for learning new patterns from the current task. This is an important trade-off which must be considered when operating in a CL scenario, and it is called \emph{stability-plasticity} trade-off \cite{wang2023comprehensive}. 
\section{Proposed training schema}
Our proposal is composed of a custom head classifier $h^t(\cdot)$ called \nhead (\nheada), and a rehearsal regularisation approach, named \nreg 
(\nrega{}). We combine these two components, producing a method capable of achieving a better stability-plasticity trade-off and leading to better results. 

\begin{figure*}[t]
\centering
\null\hfill
    \subfloat[\nreg]{
    \centering
    \includegraphics[width=0.47\linewidth]{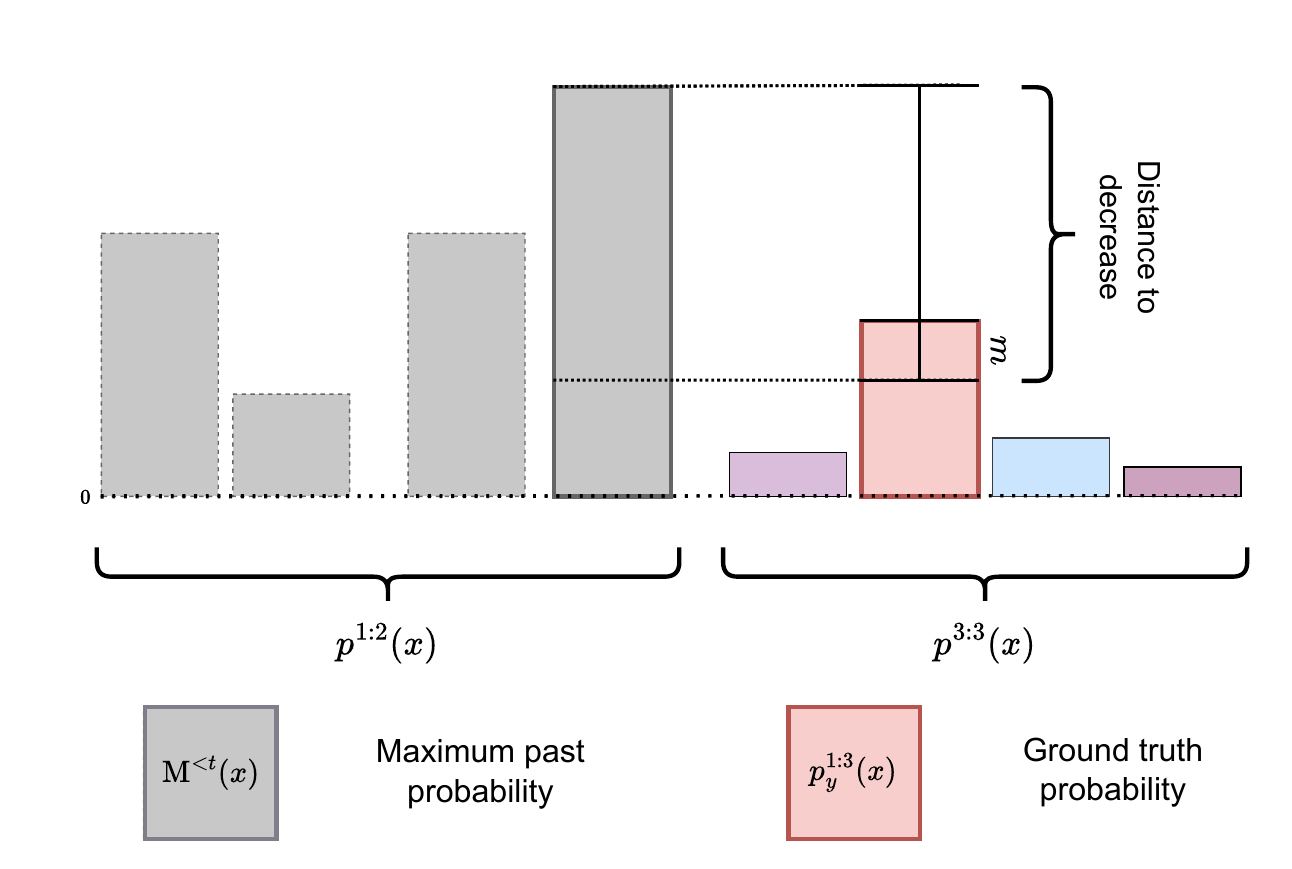}
    \label{fig:reg}
    }
    \hfill
    \subfloat[\nhead]{
    \centering
    \includegraphics[width=0.42\linewidth]{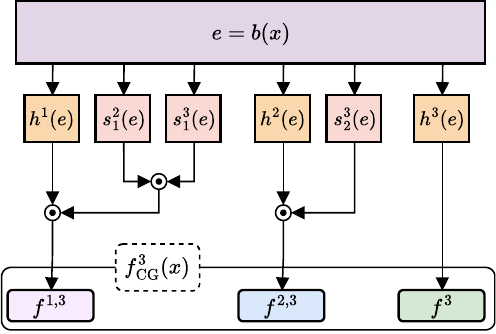}
    \label{fig:head}
    }
 \hfill\null
\caption{On the left we show the probability dampening schema (\nrega, Section \ref{sec:mbm}), in which past probabilities from past learned tasks, $p^{1:2}(x)$, are decreased up to a certain margin $m$ with respect to the ground truth probability, $p^{3:3}_y(x)$. 
On the right, we visualise the cascaded procedure (\nheada, Section \ref{sec:scaler}), which combines scaled task-wise outputs to build the final prediction $f^3_\text{CG}(x)$. Both components are visualised for a three-task scenario, in which the last one is the training one. Better viewed in colours.}
\label{fig:approach}
\end{figure*}

\subsection{\nreg (\nrega)}
\label{sec:mbm}
For a given tuple $(x, y) \in \mathcal{T}_t$, with $t>1$, the core idea of our training schema is to decrease past probabilities while training only the classes from the current task. This is done by forcing the maximum probability from past tasks to be lower than the ground truth probability, defined as the probability produced by the model for the scalar truth label y, plus a safety margin. To this end, we first denote as $p^{1:t}(x) \in \mathbb{R}^{\lvert \mathcal{Y}^{t} \rvert}$ the vector containing the probabilities calculated over all classes seen up to the task $t$ included. Using this formulation, we calculate the maximum past probability using the function $\text{M}^{<t}(x) = \max p^{1:t-1}(x)$, and the probability associated with the true training class $y$ as $p_y^{1:t}(x)$. We want the minimization procedure to satisfy the following inequality: 
%
\begin{equation}
\label{eq:constr}
p_y^{1:t}(x) - \text{M}^{<t}(x) \ge m
\end{equation}
\noindent where $m$ is the margin value which controls the safe distance between the two quantities. The constraint translates into the \nreg (\nrega) loss, defined as: 
\begin{equation}
\label{eq:margin_loss}
\text{\nrega}_t(x, y) = \max\left(0, \, \text{M}^{<t}(x) - p_y^{1:t}(x) + m \right) \\
\end{equation}
\noindent which is summed to the cross-entropy loss calculated only over classes' probability belonging to the current task $t$, $p^{t}(x) \in \mathbb{R}^{\lvert \mathrm{Y}^{t} \rvert}$ 
, with $\sum p^{t:t}_c(x) = 1$. The final loss while training on a task $t>1$ is: 
\begin{equation}
\mathcal{L}_t(x, y) = \lambda \text{\nrega}_t(x, y) + \mathcal{L}_{ce}(p^{t:t}(x), y)
\end{equation}
\noindent with $\lambda$ being a scaling value to balance the two terms, controlling the stability-plasticity trade-off. This loss works in a bidirectional manner: at the same time, the correct probability grows due to the minimisation of the cross-entropy loss calculated only over current classes, without interfering with past classes, while past probabilities are pushed down up to a certain point which is defined by the margin. Past this point, the regularisation loss becomes zero, creating a smoother and easier-to-optimise loss function. The proposed regularisation approach can be visually inspected in Figure \ref{fig:reg}.

In addition to the training loss just exposed, which is calculated only over samples from the current task, we use past samples stored in the external memory to calculate a KD regularisation term, which enforces the stability of the model. In literature, the KD is divided, usually, into General KD \cite{zhao2020maintaining}, which aggregates together logits belonging to the classes from all the previous tasks, and Task-wise approaches \cite{ahn2021ss}, which treat classes within each task separately. Based on the intuition that newly added classes are randomly produced since the corresponding classifiers are randomly initialised, resulting in logits which are lower than the already trained ones, we propose a third way. Our regularisation term operates on the output of the model as a whole, without taking into account task boundaries. By doing so, we remove the necessity of training on past samples along with the ones from the current task, preventing the rise of overfitting of such samples.  

To do that, we store a subset of past samples in an external memory $\mathcal{M}$ containing a fixed number of samples during the whole training, which is adapted by removing random portions of past samples when new ones must be stored.\footnote{When the training on a task is over, a random subset of it is saved into the memory.} The samples are drawn in a class-balanced way and are used to regularise the output of the model using the Kullback-Leibler (Kl) divergence, as $\mathcal{R}(x) = \text{KL}(p^{1:t}(x) \, \vert \vert \, \overline{p}^{1:t}(x))$, where $x \sim \mathcal{M}$, and $\overline{p}^{1:t}(x) \in \mathbb{R}^{\lvert \mathcal{Y}^t \rvert}$. The latter can be seen as a fixed teacher model with the same output as the model at the beginning of task $t$ just before starting the training process. 

\paragraph{Selection of the margin} Regarding the choice of the margin, we rely on the observation that a CIL scenario is defined in such a way that the number of classes grows with the number of tasks, which implies that the maximum probability obtainable decreases after each task. Following this intuition, the margin $m$ in Eq \eqref{eq:margin_loss} must be scaled accordingly. To do that, we use a margin value which is based on how many classes have been encountered up to the current task $t$: $m^t = \frac{1}{\lvert \mathcal{Y}^t \rvert - 1}$.
By doing so, we avoid manually selecting the margin, which could result in either a disruptive or a negligible \nrega loss when not properly done. Not only does the margin value adapt while training, but it also allows the model to self-adapt based on the complexity of the training, removing the necessity to tune an additional hyperparameter. 

\paragraph{Final loss} For a given tuple $(x, y) \in \mathcal{D}^t$ coming from the current task $t$, the final loss we propose is the following: 
\[
\mathcal{L}^{\text{\nrega}}_t(x, y,x_m) = \left\{
\begin{aligned}\quad
&\begin{aligned}
    &\displaystyle \mathcal{L}_t(x, y) + \mathcal{R}(x_m)
 \end{aligned}  & \text{if } t > 1 \\
    &  \mathcal{L}_{ce}(p^{1:1}(x),y) \quad & t = 1
 \end{aligned}
\right.
\]
\noindent in which $x_m$ are the samples drawn from the external memory $\mathcal{M}$ in a class-balanced way, $x_m \sim \mathcal{M}$, such that $\lvert x_r \rvert = \lvert x \rvert$.

\subsection{\nhead (\nheada)}
\label{sec:scaler}

As opposed to standard heads used in literature, our approach is an ensemble of small task-wise heads, which are scaled and combined to produce the final output of the model.
In this sense, our approach is a hybrid between a multi-head approach usually employed in TIL scenarios, and the single incremental one, which is the standard one used in the literature when dealing with CIL scenarios. The core idea is to help \nrega decrease past classes'  probabilities for samples coming from a new task, without operating directly on the associated classifier heads, allowing for more plasticity. To do that, past logits are scaled using additional scaling heads that operate as gating mechanisms. 

Recalling that in our scenario the forward function of our model is decoupled as $f^t(x) = h^t(e)$, with $e = b(x)$, we need to adapt our model by adding a new head each time a new task is retrieved. When a new task $t > 1$ is retrieved we firstly add a task-wise head  $h^t(e) \in \mathbb{R}^{\lvert \mathrm{Y}^t \rvert}$ into the model. Then, we create a set of scaling functions as $S^t(e) = \{s^t_i(e) \in \mathbb{R}^{\lvert \mathrm{Y}^i \rvert}\}$ for $i = 1, \dots, t-1$.
Each of these additional modules are composed of a linear layer $g^j_i(\cdot) \in \mathbb{R}^{\lvert \mathrm{Y}^t \rvert}$, followed by a scaled Sigmoid such that $s^t_i(e) = \text{Sigmoid}(\gamma \cdot g^j_i(e) + \beta)$, with $\gamma=1$ and $\beta=10$.\footnote{In the experimental section we study how these values affect the results.}
Scaling the logits simplifies past classes' regularisation without interfering directly with past heads. Mathematically, given the current training task index $t$ and a past task $i<t$, the output associated to $i$ is calculated as:
\begin{align}
\label{eq:scaled_function}
f^{i, t}(x) = f^i(x) \cdot \prod_{j=i+1}^{t}s^j_i(b(x)) = h^i(b(x))  \cdot \prod_{j=i+1}^{t} \text{Sigmoid}(\gamma \cdot g^j_i(b(x)) + \beta)
\end{align}
The final output of the model is produced by concatenating all such vectors, including the unscaled one from the current training class: 
\begin{equation}
f^{t}_{\text{\nheada}}(x) = [f^{1, t}(x), f^{2, t}(x), \dots,  f^t(x)] \in \mathbb{R}^{\rvert \mathcal{Y}^t \lvert}
\end{equation}
%
\noindent where $t$ is the last task seen during the training procedure. The proposed head classifier can be visually inspected in Figure \ref{fig:head}.

\section{Experimental analysis}

In this section, we define the experimental setup and show the results against well-established baselines. Then, we proceed with an in-depth analysis of our proposal and its components. Further experiments and analysis can be found in Appendix \ref{app:additional_experiments}. The overall code is based on the Avalanche library \cite{lomonaco2021avalanche} and it is available on the official repository\footnote{ \href{https://github.com/jaryP/CIL-Margin-Dampening-Gated-Classifier}{https://github.com/jaryP/CIL-Margin-Dampening-Gated-Classifier}}.

\subsection{Experimental settings}
%

%

\paragraph{Benchmarks} 
To evaluate our proposal, we use a variety of established CL benchmarks. Each one is built from a vision classification dataset, which is split into $\text{T}$ disjoint sets, each one containing $\frac{C}{\text{T}}$ classes, where $C$ is the number of classes in the original dataset.\footnote{These subsets follow the rules exposed in Section \ref{sec:cl_definition}.} Usually, in literature, the classes are grouped incrementally. However, how the classes are divided is crucial and can radically change the obtained results. To cover a wider spectrum of possible scenarios, as well as study the stability of each CL approach, we group the classes randomly each time a scenario is built, resulting in a different scenario with its complexity. 
We evaluate our proposal on three different datasets, used to create scenarios having growing difficulty, which are: \textbf{CIFAR10} contains 10000 $32 \times 32$ sized images, divided into 10 classes. Using this dataset we build the C10-5 scenario, in which we have 5 tasks, each one having 2 classes (\textbf{C10-5}); \textbf{CIFAR100} has the number and size of the images are the same as in CIFAR10, but we have 100 classes, resulting in fewer samples per class, which we use to create a scenario with 10 tasks (\textbf{C100-10})
; and \textbf{TinyImageNet} (\textbf{TyM}), which is a subset of 200 classes from ImageNet \cite{imagenet} which contains $64 \times 64$ sized images. Each class has 500 training and 50 testing images, and we used this dataset to build scenarios in which each task contains 20, 10 and 5 classes, respectively \textbf{TyM-10},  \textbf{TyM-20}, and \textbf{TyM-50}. Using such scenarios we have a wide range of difficulties, helping us understand how our approach behaves with respect to the baselines.  





\paragraph{Architectures and training details}
We use a ResNet20 model \cite{he2016deep} trained from scratch for CIFAR scenarios, while we used a ResNet18 for the others. To have easy-to-read comparisons, we trained all the models using the same number of epochs, batch size, and SGD optimizer with a learning rate of $0.01$ and a momentum of $0.8$. For C10-5 we train each task for 20 epochs, using a batch size of 32. We train for the same number of epochs also for C100-10, but with a batch size of 16. Regarding TyM scenarios, we train the model for 30 epochs with a batch size of 32. Moreover, we use the same augmentation schema for each dataset (a random cropping, followed by a horizontal flipping with a probability of $0.5$ and the normalization step), and no scheduling schemes or further regularisation methods are used (if not explicitly needed by a CL method).

For each combination of scenario and memory size, we run 5 experiments, each time incrementally setting the random seed (from 0 to 4), resulting, for each approach, in the same starting model and the same scenario, built up by randomly grouping the classes.  

\paragraph{Hyperparameters}
For each CL approach, we select the best set of hyperparameters through a grid search. The approaches are evaluated over a portion of the training data split (10\%) which is used only for evaluation purposes. The hyperparameters giving the best results on this split, after training on the train set in a CIL scenario, are the ones used in the final experimental evaluation. We do not rely on the hyperparameters selected in each paper to have better comparable results, all obtained under the same unified experimental environment. Each combination of approach, memory size, and scenario is evaluated once. We evaluated each model over a single shared seed. The evaluated hyperparameters for each approach, as well as the best ones, can be found in Appendix \ref{ap:hps}. 



\paragraph{Metrics}
To evaluate the efficiency of a CL method, we use two widely-used metrics \cite{diaz2018don}. The first one, called Scenario Accuracy (ACC), measures the final accuracy obtained across all the tasks' test splits, while the second one, called Backward Transfer (BWT), tells us how much of that past accuracy is lost during the training on upcoming tasks; both metrics are averaged over all tasks once the training on all of them is over. 
Appendix \ref{ap:metrics} contains the details on how such metrics are calculated and why both are important.
\begin{table*}[t]
\centering
\setlength{\extrarowheight}{0pt}
\addtolength{\extrarowheight}{\aboverulesep}
\addtolength{\extrarowheight}{\belowrulesep}
\setlength{\aboverulesep}{0pt}
\setlength{\belowrulesep}{0pt}
\resizebox{0.8\linewidth}{!}{%
\begin{tabular}{l|cccc|cccc|} 
\cline{2-9}
   & \multicolumn{4}{c|}{C10-5}   & \multicolumn{4}{c|}{C100-10} \\ \hline   \hline
   
\rowcolor[rgb]{0.933,0.933,0.925} \multicolumn{1}{|c|}{Naive} & \multicolumn{4}{c|}{\result{18.62}{0.1}{96.25}{1.0}}  & \multicolumn{4}{c|}{\result{8.0}{0.3}{82.4}{0.1}}   \\

\multicolumn{1}{|c|}{Cumulative}  & \multicolumn{4}{c|}{ \result{86.8}{0.5}{2.7}{1.1}}   & \multicolumn{4}{c|}{\result{61.4}{0.4}{3.4}{0.4}}  \\ 

\hline\hline
\multicolumn{1}{c}{Memory Size} & \multicolumn{1}{c}{200} & \multicolumn{1}{c}{500} & \multicolumn{1}{c}{1K} & \multicolumn{1}{c}{2K} & \multicolumn{1}{c}{200} & \multicolumn{1}{c}{500} & \multicolumn{1}{c}{1K} & \multicolumn{1}{c}{2K} \\ 
\hline\hline

\rowcolor[rgb]{0.933,0.933,0.925} \multicolumn{1}{|c|}{Replay} & \result{38.8}{3.3}{69.4}{5.0} & \result{50.2}{3.6}{53.6}{5.1} & \result{61.1}{2.3}{38.8}{4.3} & \result{61.1}{2.3}{31.0}{4.0} & \result{12.6}{0.5}{74.0}{1.2} & \result{17.6}{0.4}{66.6}{1.8} & \result{25.8}{0.7}{56.1}{1.6} & \result{32.6}{1.2}{46.1}{0.5} \\

\multicolumn{1}{|c|}{GEM \cite{lopez2017gradient}}  & \result{18.8}{0.1}{88.9}{5.2} & \result{20.8}{0.2}{89.7}{2.7} & \result{21.8}{2.7}{82.2}{7.0} & \result{22.3}{1.9}{82.9}{7.1} & \result{17.6}{1.1}{68.1}{1.6} & \result{18.4}{2.4}{59.1}{5.6} & \result{19.1}{1.7}{54.3}{3.9} & {--} \\

\rowcolor[rgb]{0.933,0.933,0.925} \multicolumn{1}{|c|}{DER \cite{buzzega2020dark}}  & \result{43.1}{4.2}{64.4}{6.3} & \result{55.1}{3.5}{48.4}{5.3} & \cellcolor{purple!15} \result{63.7}{3.5}{36.4}{5.7} & \cellcolor{purple!15} \result{70.4}{1.6}{26.1}{3.2} & \result{12.7}{0.7}{74.7}{1.1} & \result{19.8}{0.6}{64.7}{0.8} & \result{27.4}{0.6}{54.9}{1.1} & \result{36.4}{0.3}{43.2}{0.8} 
  \\ 

\multicolumn{1}{|c|}{GDUMB \cite{prabhu2020gdumb}} &  \result{24.5}{1.0}{24.2}{5.2} & \result{31.1}{1.6}{23.5}{7.3} & \result{36.8}{1.9}{26.2}{8.2} & \result{45.2}{1.6}{25.0}{3.1} &   \result{3.6}{0.4}{47.8}{6.4}  & \result{6.7}{0.2}{28.6}{2.0} & \result{9.5}{0.4}{14.7}{0.7} &\result{14.1}{0.4}{17.4}{2.0}  \\

\rowcolor[rgb]{0.933,0.933,0.925} \multicolumn{1}{|c|}{RPC \cite{pernici2021class}} & \result{38.0}{2.8}{70.2}{4.6} & \result{50.4}{5.4}{53.1}{7.7} & \result{58.4}{1.9}{42.2}{2.25} &  \cellcolor{purple!15} \result{69.6}{2.5}{24.9}{4.1} & \result{12.6}{0.6}{73.8}{1.4} & \result{18.5}{0.8}{65.8}{1.3} & \result{25.2}{1.0}{56.7}{0.9} & \result{32.2}{1.5}{46.6}{0.8} \\

\multicolumn{1}{|c|}{SS-IL \cite{ahn2021ss}}  & \result{41.2}{1.2}{25.4}{2.0} & \result{45.6}{1.1}{21.4}{3.3}  & \result{47.3}{0.9}{21.2}{2.0} & \result{52.1}{2.0}{14.8}{1.8} & \cellcolor{purple!15} \result{19.3}{1.0}{34.1}{4.5} & \result{24.2}{0.6}{27.3}{2.7} & \result{27.5}{0.8}{20.7}{1.8} & \result{28.8}{0.9}{16.6}{1.1} \\

\rowcolor[rgb]{0.933,0.933,0.925} \multicolumn{1}{|c|}{ER-ACE \cite{caccia2022new}} &  \result{51.5}{1.1}{22.1}{1.8}  &  \cellcolor{purple!15}  \result{57.5}{1.6}{17.5}{3.2} & \cellcolor{purple!15} \result{63.4}{0.8}{11.2}{1.7} &   \result{67.5}{1.3}{6.7}{1.4} & \cellcolor{purple!15} \result{19.4}{0.3}{45.0}{0.9} & \result{26.1}{0.6}{37.3}{1.4} &  \result{31.2}{0.2}{30.9}{1.2} &  \result{36.3}{0.7}{24.1}{1.6} \\

\multicolumn{1}{|c|}{ER-LODE \cite{liang2023loss}}  & \result{49.3}{5.6}{14.4}{3.2} & \resultp{56.9}{1.5}{4.4}{3.7} & \resultp{58.5}{1.2}{13.1}{5.3}  & \resultp{60.4}{2.4}{17.0}{4.7} & \cellcolor{purple!15} \result{19.4}{1.0}{40.6}{2.2} &  \result{28.5}{0.5}{33.7}{2.2} & \cellcolor{purple!15} \result{33.3}{0.5}{20.4}{2.4} & \cellcolor{purple!15} \result{38.2}{0.9}{19.0}{1.4}
 \\ \hline \hline

\rowcolor[rgb]{0.933,0.933,0.925} \multicolumn{1}{|c|}{LD} & \cellcolor{purple!15} \result{55.9}{3.7}{36.8}{9.1} & \cellcolor{purple!15} \result{58.9}{1.7}{32.4}{4.3} & \result{57.9}{3.7}{33.4}{5.4} & \result{55.9}{3.7}{32.5}{5.9} & \cellcolor{purple!15} \result{19.6}{3.1}{50.6}{11.2} & \cellcolor{purple!15} \result{29.6}{2.8}{30.7}{13.25} &  \result{30.1}{2.5}{27.0}{12.7} & \result{35.1}{1.1}{19.3}{1.6}  \\

 \multicolumn{1}{|c|}{\napproach (ours)} & \cellcolor{blue!15} \result{61.4}{1.9}{27.9}{2.3} & \cellcolor{blue!15} \result{65.5}{3.2}{35.7}{5.7} &  \cellcolor{blue!15} \result{69.4}{1.2}{28.3}{2.3} & \cellcolor{blue!15} \result{75.68}{0.7}{17.2}{2.3} & \cellcolor{blue!15} \result{21.8}{0.8}{53.9}{1.7} & \cellcolor{blue!15} \result{30.3}{0.7}{41.1}{1.6} & \cellcolor{blue!15} \result{35.6}{0.5}{21.3}{1.1} &\cellcolor{blue!15}  \result{41.3}{1.1}{25.7}{0.7} \\
\bottomrule
\end{tabular}
}
\caption{The results obtained for each scenario, formatted as $\text{ACC} (\uparrow) \vert \text{BWT} (\downarrow)$ averaged over 5 runs; the standard deviations are also shown. The best results for each combination of scenario and memory are \mybox[fill=blue!15]{highlighted}. For a better comparison, the second best results are also \mybox[fill=purple!15]{highlighted}. Better viewed in colors.}
\label{table:results}
\end{table*}

\paragraph{Baselines}
For a fair comparison we select only CL algorithms for which we can explicitly control the size of the external memory. The selected ones are: \textbf{Experience Replay} \cite{chaudhry2019tiny, buzzega2021rethinking}, \textbf{Greedy Sampler and Dumb learner} (GDumb) \cite{prabhu2020gdumb}, \textbf{Experience Replay with Asymmetric Cross-Entropy} (ER-ACE) \cite{caccia2022new}, \textbf{Separated Softmax for Incremental Learning} (SS-IL) \cite{ahn2021ss}, \textbf{Gradient Episodic Memory} (GEM) \cite{lopez2017gradient}, \textbf{Dark experience replay} (DER) \cite{buzzega2020dark}, in its DER++ version, \textbf{Regular Polytope Classifier} (RPC) \cite{pernici2021class}, and \textbf{Rehearsal Loss Decoupling} (ER-LODE) \cite{liang2023loss}. 
In addition to such methods, we also propose a simple baseline called \textbf{Logits Distillation} (LD), in which an external class balanced memory contains images used to regularize the logits with the KL divergence (weighted with a constant $\alpha$), without involving any classification loss over past samples. To do so, a copy of the model, used as the teacher model, is created at the beginning of the training and used to regularise the whole output, as in our main proposal \nreg (Section \ref{sec:mbm}). This baseline is useful for studying the effect of regularising the output without involving any classification loss over past samples. 
In addition, we also used \textbf{Naive} as the lower bound, in which the model is trained sequentially without any CF mitigation technique, and the upper bound called \textbf{Cumulative}, which trains the models on all the tasks up to the current one combined into a single training dataset.
A more detailed overview of each baseline approach can be found in Appendix \ref{ap:baselines}. 

\subsection{Experimental results}
\subsubsection{Accuracy and forgetting}
\begin{table*}[t]
\centering
\setlength{\extrarowheight}{0pt}
\addtolength{\extrarowheight}{\aboverulesep}
\addtolength{\extrarowheight}{\belowrulesep}
\setlength{\aboverulesep}{0pt}
\setlength{\belowrulesep}{0pt}
\resizebox{0.7\linewidth}{!}{%
\begin{tabular}{l|cc|cc|cc|} 
\cline{2-7}
& \multicolumn{2}{c|}{TyM-10}   & \multicolumn{2}{c|}{TyM-20}   & \multicolumn{2}{c|}{TyM-50} \\ \hline 
   
\rowcolor[rgb]{0.933,0.933,0.925} \multicolumn{1}{|c|}{Naive} & \multicolumn{2}{c|}{\result{6.67}{0.2}{66.28
}{1.1}}  & \multicolumn{2}{c|}{\result{3.86}{0.5}{74.6}{0.1}}  & \multicolumn{2}{c|}{\result{1.7}{0.2}{83.9}{0.8}}   \\

\multicolumn{1}{|c|}{Cumulative}  & \multicolumn{2}{c|}{ \result{35.8}{0.2}{12.6}{0.3}}   & \multicolumn{2}{c|}{\result{33.52}{0.1}{14.2}{0.2}} & \multicolumn{2}{c|}{\result{33.2}{0.3}{16.9}{0.8}}  \\ 

\hline\hline
\multicolumn{1}{c}{Memory Size} & \multicolumn{1}{c}{2000} & \multicolumn{1}{c}{5000} & \multicolumn{1}{c}{2000} & \multicolumn{1}{c}{5000} & \multicolumn{1}{c}{2000} & \multicolumn{1}{c}{5000}  \\ 
\hline\hline

\rowcolor[rgb]{0.933,0.933,0.925} \multicolumn{1}{|c|}{Replay} & \result{12.5}{0.8}{55.9}{1.1} & \result{18.3}{0.2}{46.9}{0.8} & \result{8.12}{0.3}{63.1}{0.7} & \result{13.2}{0.3}{54.2}{0.6} & \result{6.5
}{0.2}{73.2}{0.8} & \result{12.1}{0.4}{63.1}{0.6} \\

\multicolumn{1}{|c|}{DER \cite{buzzega2020dark}}  & \result{16.2}{1.2}{54.4}{1.1} & \result{22.9}{0.8}{43.2}{0.6} &  \result{11.3}{0.9}{43.5}{0.8} &  \result{14.2}{0.6}{38.4}{0.5} & \result{10.1}{0.4}{33.2}{0.6} & \result{12.6}{0.7}{30.6}{0.6} \\ 

\rowcolor[rgb]{0.933,0.933,0.925} \multicolumn{1}{|c|}{ER-ACE \cite{caccia2022new}} &  \cellcolor{purple!15} \result{21.2}{0.2}{27.9}{0.4}  &   \cellcolor{purple!15} \result{24.3}{0.2}{24.5}{0.1} &  \result{14.1}{0.8}{26.2}{0.7} &   \result{17.7}{0.5}{24.9}{0.7} & \result{11.5}{0.3}{23.5}{0.8} & \result{15.8}{0.6}{23.6}{0.5}  \\

\multicolumn{1}{|c|}{ER-LODE \cite{liang2023loss}}  & \cellcolor{purple!15} \result{20.1}{0.3}{34.6}{0.2} & \cellcolor{purple!15} \result{23.9}{0.3}{25.8}{0.2} & \cellcolor{purple!15} \result{17.8}{0.6}{28.6}{0.6} & \cellcolor{purple!15}  \result{21.9}{0.4}{22.3}{0.6} & \cellcolor{purple!15}\result{13.5}{0.3}{25.6}{0.2}  & \cellcolor{purple!15} \result{17.2}{0.3}{10.5}{0.2} \\ \hline \hline


 \rowcolor[rgb]{0.933,0.933,0.925}  \multicolumn{1}{|c|}{\napproach (ours)} & \cellcolor{blue!15} \result{24.2}{0.2}{23.6}{0.3}     &  \cellcolor{blue!15}\result{27.5}{0.3}{11.9}{0.4} & \cellcolor{blue!15} \result{21.2}{0.6}{27.2}{0.7} & \cellcolor{blue!15} \result{25.3}{0.3}{14.7}{0.6} & \cellcolor{blue!15} \result{15.4}{0.3}{32.2}{0.4} & \cellcolor{blue!15} \result{21.6}{0.6}{19.3}{0.8} \\
\bottomrule
\end{tabular}
}
\caption{The results obtained for each scenario, formatted as $\text{ACC} (\uparrow) \vert \text{BWT} (\downarrow)$ averaged over 5 runs; the standard deviations are also shown. The best results for each combination of scenario and memory are \mybox[fill=blue!15]{highlighted}. For a better comparison, the second best results are also \mybox[fill=purple!15]{highlighted}. Only the best CL approaches are shown. Better viewed in colors.}
\label{table:tyn_results}
\end{table*}
Table \ref{table:results} shows the results of CIFAR-like scenarios. By examining it we can observe that the simple Replay approach often achieves higher scores than more elaborated ones. It is clear when comparing Replay and GEM: the first reaches, on average, higher results than the latter, even when the BWT is worst. Moreover, it scales better with the memory size, while GEM seems to saturate once a limit is reached .\footnote{Some GEM results are omitted due to the huge amount of time required to complete a training procedure.} This is also true if we compare Replay against methods presented as good to fight the class imbalance issue arising from the smaller size of the memory if compared to the current training dataset, such as SS-IL and RPC. Such methods also saturate the results when the memory grows, due to the focus on stability achieved by imposing strong ergularisation terms, overshadowing the plasticity. This combination leads to better BWT but lower accuracy.  Prioritising stability could work when the number of tasks is limited, but its capacity to produce competitive results is negatively influenced when the number of tasks increases. DER, which simultaneously regularises the logits and trains over past samples, achieves better results than the already-analysed approaches. However, its performances drop when the scenario is harder. 

Table \ref{table:tyn_results}, which contains only the best baselines, shows the results obtained on multiple TyM scenarios. As before, Replay struggles to achieve good results, as well as DER, which fails when the number of tasks increases. The other two approaches, ER-LODE and ER-ECE, are the best baselines also for CIFAR-like scenarios. However, ER-ACE struggles to achieve competitive results when the number of tasks increases, while ER-LODE scales better. 

Our proposal constantly reaches better results on all the scenarios tested. When the scenario is easy, such as C-10, it overcomes AR-ACE by 10 percentage points when using a small memory (200) and, even if the gap is not preserved over all the experiments, the results are constantly better and scale better with the external memory's size, showing that our approach achieves a better stability-plasticity trade off. It achieves better results also when dealing with a large number of tasks (such as TyM-50), showing its adaptability. To further understand how our proposal behaves, we proceed to an in-depth analysis of the components of the approach.

\subsubsection{Stability-plasticity trade-off}
\begin{figure}[t]
\centering
    \subfloat{
    \includegraphics[width=0.35\linewidth]{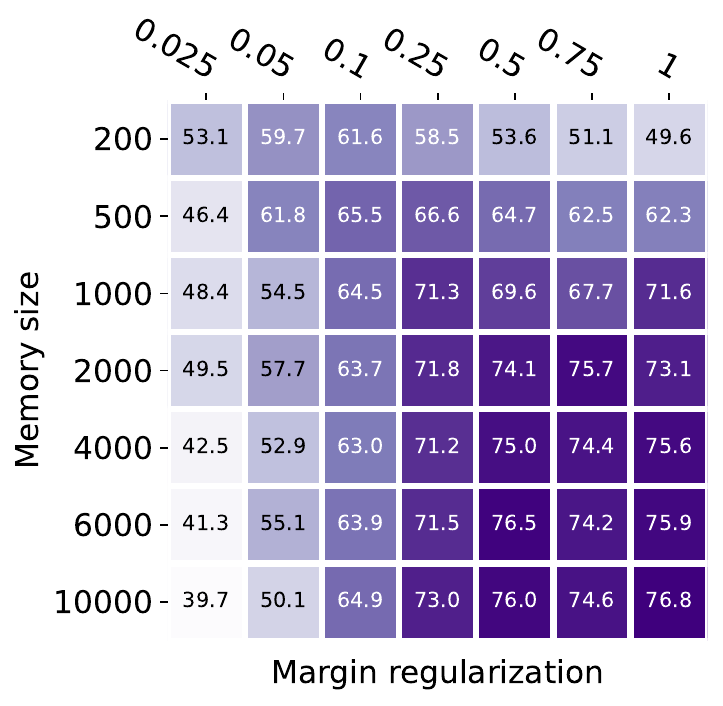}
    \label{fig:tradeoff_heatmaps_1}
    }
    \subfloat{
    \includegraphics[width=0.35\linewidth]{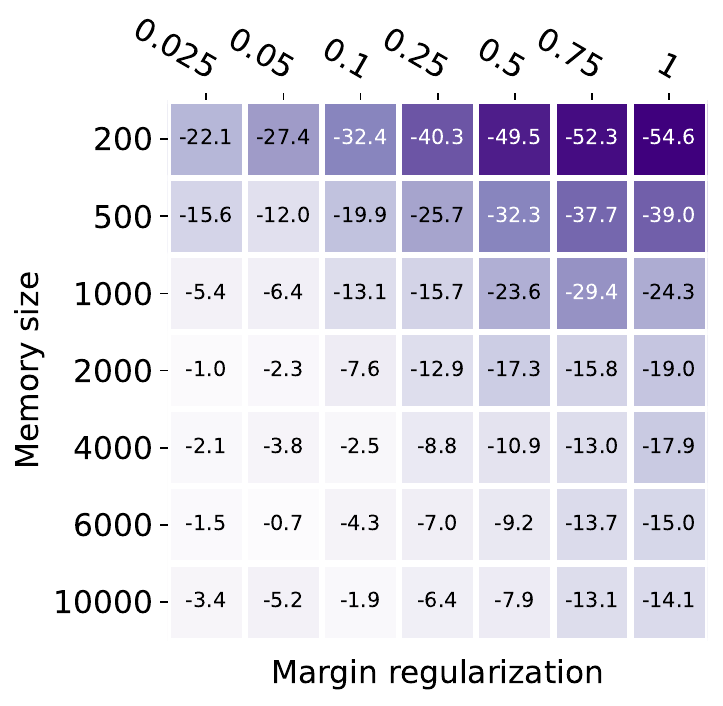}
    \label{fig:tradeoff_heatmas_2}
    }
\caption{How the accuracy (left) and the BTW (right) scores are affected when varying the memory and the past margin regularisation. The results are obtained on ResNet-20 trained on C10-5.}
    \label{fig:tradeoff_heatmaps}
\end{figure}
Here, we analyse how our approach controls the trade-off between plasticity and stability. The only parameter of our regularisation schema is $\lambda$, which is combined with the memory's size to achieve the desired trade-off. Figure \ref{fig:tradeoff_heatmaps} shows the results of training a ResNet-20 on the C10-5 scenario. It shows that these values can be combined to achieve the best stability-plasticity trade-off also when operating with sample memories. For example, combining a small memory (e.g. 200) with a high regularisation term (higher than 0.1) leads to bad results, since the memory does not contain enough samples to balance the forgetting. However, better results can be achieved when the regularisation term is $0.1$ or lower, giving the best results for such a combination of memory size and the regularisation term, and showing that a good trade-off can be easily obtained. 
These findings suggest that increasing the memory's size allows for a stronger regularisation term, and such a combination improves the results. On the other hand, having a low regularisation term (e.g.,  0.01) leads to a training schema focused on the stability of the model, which is incapable of properly learning current classes when using any memory size. Such combinations lead to a low BWT and low accuracy, symptoms of a focus on stability preservation.

In the end, such results show that, as opposed to other approaches, the choice of the regularisation term $\lambda$ and the memory size affect the achievable results predictably. These two quantities are easy to balance, resulting in better results overall.
\subsubsection{Memory overhead}
\begin{figure}[t]
    \subfloat[C10-5 scenario.]{
    \includegraphics[width=0.32\linewidth]{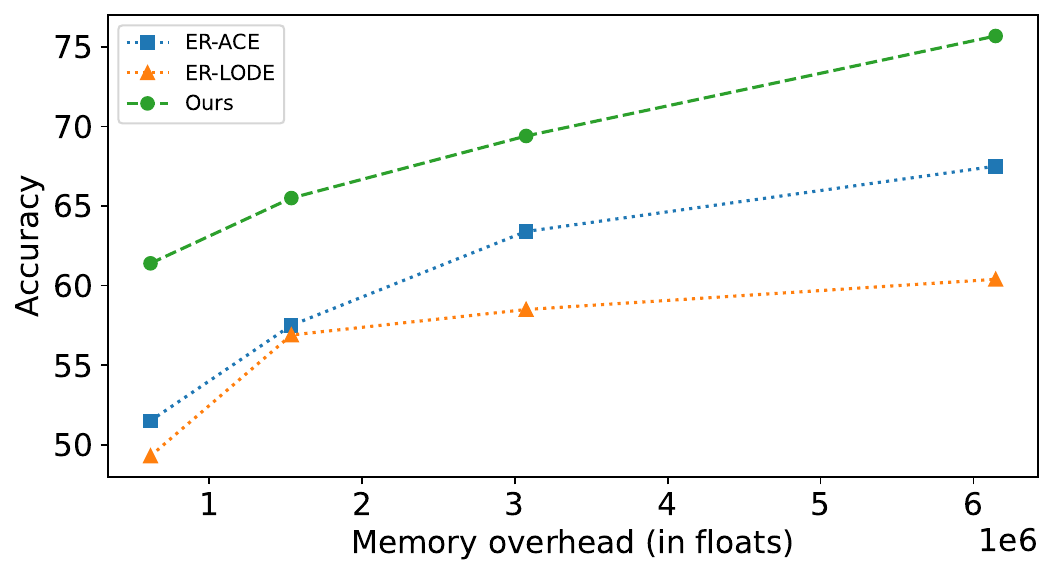}
    \label{fig:floats_1}
    }
    \hfill
    \subfloat[C100-10 scenario.]{
    \includegraphics[width=0.32\linewidth]{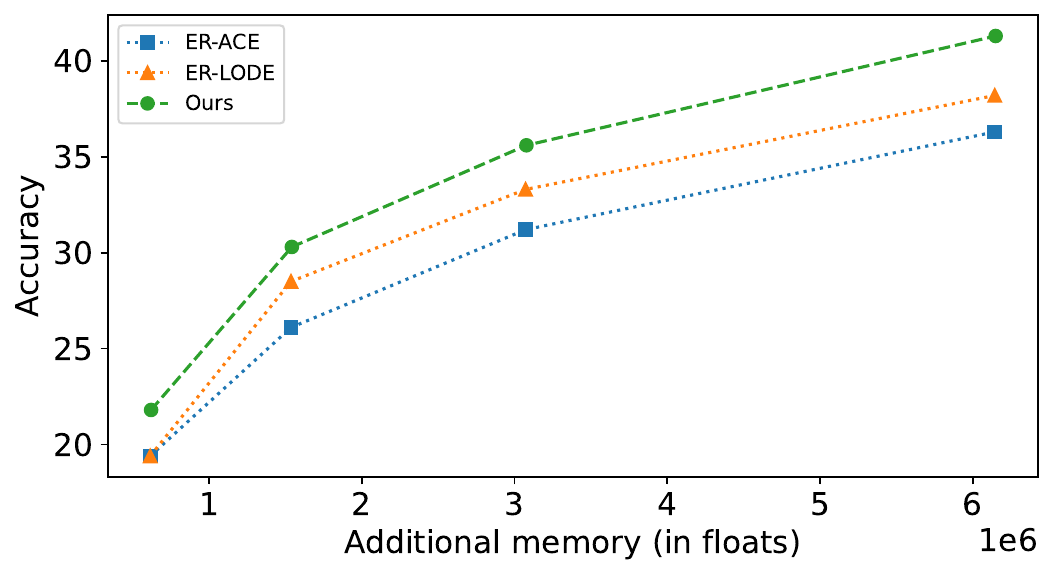}
    \label{fig:floats_2}
    }
    \hfill
    \subfloat[Tym-20 scenario.]{
    \includegraphics[width=0.32\linewidth]{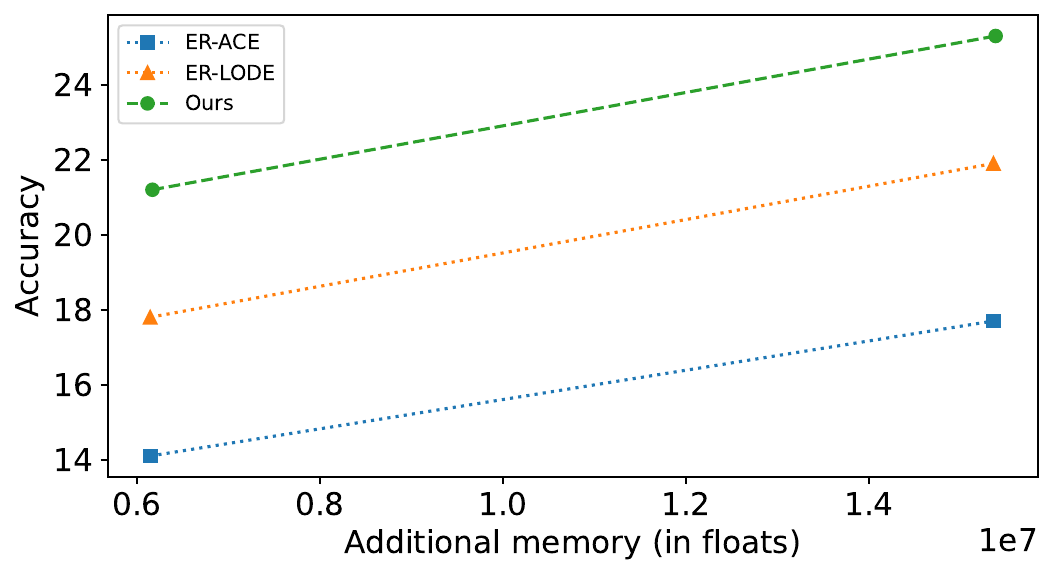}
    \label{fig:floats_3}
    }
    \caption{The additional floats required by the approaches against the achieved accuracy. As additional floats, we count all the pixels in the memory, as well as additional parameters that a method requires. 
    }
     \label{fig:floats}
\end{figure}
Here, we compare the required additional memory of our proposal with other baselines. To do that, we count the number of additional floats, counted as the total number of pixels for the images in the memory plus parameters, that a method requires to mitigate CF compared to the model used in a Naive training approach, in which no memory nor additional parameters are used
; the details on how such floats are calculated are present in Appendix \ref{ap:overhead}. The results are in Figure \ref{fig:floats}, which shows the required floats against the achieved accuracy for multiple methods. The images show that our approach achieves better results when using the same number of additional floats than ER-ACE and ER-LODE. This happens because the additional modules in \nheada require a negligible memory overhead compared to the rehearsal memory for the tested architectures. Despite the number of heads added by our approach growing quadratically (see Appendix \ref{app:time} for further details), the overall number of additional parameters is contained if compared to the dimension of the external memory. 

\subsubsection{\nhead ablations}
%
%
\begin{table*}[t]
\centering
\resizebox{0.8\linewidth}{!}{%
\begin{tabular}{r|cccc|cccc|cc|} 
\cline{2-11}
   & \multicolumn{4}{c|}{C10-5}   & \multicolumn{4}{c|}{C100-10} & \multicolumn{2}{c|}{TyM-20} \\ \hline   \hline

\multicolumn{1}{c}{Memory Size} & \multicolumn{1}{c}{200} & \multicolumn{1}{c}{500} & \multicolumn{1}{c}{1K} & \multicolumn{1}{c}{2K} & \multicolumn{1}{c}{200} & \multicolumn{1}{c}{500} & \multicolumn{1}{c}{1K} & \multicolumn{1}{c}{2K} & \multicolumn{1}{c}{2K} & \multicolumn{1}{c}{5K} \\ 
\hline\hline

\rowcolor[rgb]{0.933,0.933,0.925} \multicolumn{1}{|c|}{\napproach (ours)} & \cellcolor{blue!15} 61.4 & \cellcolor{blue!15} 65.5 & \cellcolor{blue!15} 69.4 & \cellcolor{blue!15} 75.6 & \cellcolor{blue!15} 21.8 & \cellcolor{blue!15} 30.3 & \cellcolor{blue!15} 35.6 & \cellcolor{blue!15} 41.3 & \cellcolor{blue!15} 21.2 & \cellcolor{blue!15} 25.3 \\

\multicolumn{1}{|l|}{$\hookrightarrow$ S-\nheada} & 58.0 & 62.9 & 66.9 & 69.4 & 19.3 & 27.7 & 34.7 &  38.7 & \cellcolor{blue!15} 20.2 & \cellcolor{blue!15} 25.1 \\

\multicolumn{1}{|l|}{$\hookrightarrow$ No \nheada} & 55.4 & 62.7 & 65.3 & 68.9 & \cellcolor{blue!15} 21.00 & 27.82 &  34.7 &  36.9 & 19.2 & 23.1 \\ \hline \hline

\rowcolor[rgb]{0.933,0.933,0.925}  \multicolumn{1}{|l|}{ER-ACE} & 50.3 &  56.3 & 62.5 & 67.1 & 19.5 & 26.2 & 31.1 &  36.3 & 14.1 & 17.7 \\

\multicolumn{1}{|l|}{
$\hookrightarrow$ \nheada} & 52.7 & 61.3 & 63.1 & 68.3 & \cellcolor{blue!15} 21.4 & 28.2
 & 32.4 & 35.0 & 16.4 & 20.3 \\ \hline



\end{tabular}
}
\caption{
The accuracies obtained on various scenarios when varying the components of the classifier layer. \napproach is our main proposal, while S-\nheada
and No \nrega are, respectively, our approach when a single scaling head is used instead of the whole scaling architecture $S^t$, and when \nheada is not used at all. ER-ACE is the method we already introduced, which we also couple with the head we proposed (\nheada). The best results for each combination of scenario and memory are \mybox[fill=blue!15]{highlighted}.}
\label{table:ablation}
\end{table*}
In this section, we analyse how much the proposed classifier \nheada affects the results. To this end, we compare our proposal when the scaling functions in the set $S^t(x)$ are used (as in the main experiments) or not (which is, basically, an incremental classifier). Additionally, we use also a version of \nheada in which the scaling head is just one and no cascaded component is used. Moreover, to understand if \nheada could improve other approaches, we also compare the results achieved by ER-ACE when using or not the \nheada classifier. 

Table \ref{table:ablation} contains the results of both experiments. Looking at the ER-ACE results, we can see that our approach is capable of marginally improving the accuracy with respect to the counterpart that uses the incremental head classifier, especially when the memory size is contained; however, this improvement diminishes or disappears when its dimension grows. Regarding our approach, the results are worse when the logits are not scaled, regardless of the memory size. However, the results are always better than ER-ACE. Instead, when using a single scaling head, the achieved results are competitive but lower than the ones obtained using multiple scaling heads. Intuitively, this happens because the scaling approach helps to mitigate the class unbalancing issue by giving the training procedure two ways to decrease past logits: by directly decreasing them or by decreasing the scaling value. This improves the plasticity without negatively affecting the stability, leading to better results. These aspects make our proposal competitive over all the benchmarks selected, with all possible combinations of memory size. 

\section{Limitations and conclusion}
We proposed a novel rehearsal-regularisation approach which combines a constraint-based regularisation schema and a scaled classifier head, which builds the final prediction vector using a cascaded approach. Combining these components creates a CL method which achieves better results than the compared baselines. Our approach takes advantage of a soft constraint, which allows for smoother regularisation and less forgetting, even when an external memory contains few samples per class. We also extensively evaluated our proposal, to understand how and why the components affect the results. Our approach has no drawback when it comes to accuracy and forgetting, nor even when it comes to memory overhead. However, due to the nature of the training head, the number of additional heads grows quadratically with the number of tasks. 

In the future, we will delve more into the theoretical analysis of rehearsal samples overfitting in CL, which we empirically observed and shown in the paper. We will also extend the approach for CL scenarios in which the task boundaries are not well defined, as well as Online CL scenarios. Moreover, we want to experiment with more classification heads, to reduce the time as well as the memory complexity of our proposal.



\bibliography{main}
\bibliographystyle{abbrvnat}

\newpage
\appendix

\section{Regularising and learning on past samples}
\label{ap:reg}
\begin{wrapfigure}{R}{0.4\textwidth}
  \centering
\includegraphics[width=0.4\columnwidth]{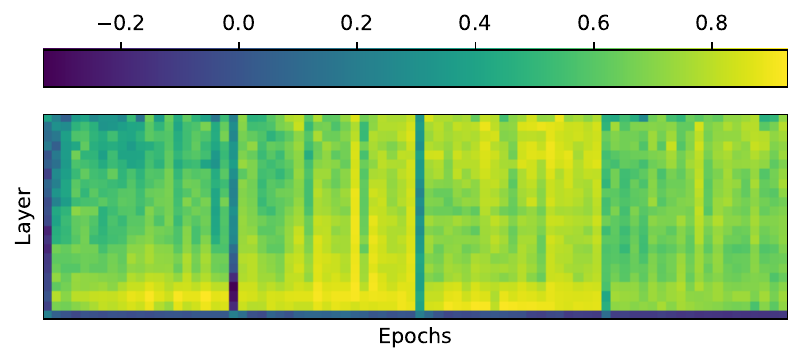}
  \caption{The cosine distance between gradients produced by the loss calculated on rehearsal samples and the ones related to the Knowledge Distillation regularisation. The last layer (last row of the heatmap) is the one which constantly has the two gradients pointing toward different directions.}
  \label{fig:der_reg}
\end{wrapfigure}
We hypothesize that a hybrid approach which, at the same time, uses rehearsal samples as training samples as well as regularisation ones is sub-optimal. Figure \ref{fig:der_reg} shows how the gradients diverge when using DER \cite{buzzega2020dark}, when training Resnet-20 on C10-5, with a memory having a size of 500. The image shows the difference, calculated over the output channels, between the gradients obtained using the standard cross entropy and the one obtained for the knowledge distillation regularisation, both calculated over samples in the memory.

We can see that the method produces classification layer's gradients giving negative similarity, suggesting that the two terms of the training loss try to move the model towards two different directions: in the first one the model satisfies the cross entropy (with the risk of overfitting over past samples), while the second one tries to keep the logits fixed. Even if competitive results are achievable by carefully balancing the loss terms, we advocate that the final results are highly sub-optimal.

\section{Related work}
\label{ap:rw}
Existing CL approaches can be mainly categorised into three categories, even if, most of the time, an approach can belong to multiple categories at the same time. Regularisation approaches \cite{Kirkpatrick_2017, zenke2017continual, Aljundi_2018, ahn2021ss, chaudhry2021using, frascaroli2023casper, gomez2022continually} introduce additional terms in the loss function to force the model to preserve knowledge that is crucial to keep solving past learned tasks while learning to solve newer ones. Architectural approaches \cite{rusu2016progressive, mallya2018packnet, golkar2019continual, douillard2020podnet, POMPONI2021407, zhai2023masked, wortsman2020supermasks} directly operate on the model itself, by isolating weights or dynamically expanding the model capacity. Rehearsal-based approaches implement an external memory containing examples from prior tasks and usually train the model jointly with the current task \cite{buzzega2020dark, chaudhry2019tiny, prabhu2020gdumb, pham2023continual, wang2022sparcl, xu2023continual, bang2021rainbow, POMPONI2021407}. 
A subset of such methods is called Pseudo-rehearsal, in which a generative model takes the place of the memory \cite{van2020brain, lao2020continuous, jimaging8040093, 10.5555/3618408.3618842, pomponi2023continual}. Although generative models are susceptible to issues like forgetting and mode collapse, they overcome the lack of diversity present in bounded memory buffers. 

\section{Baselines}
\label{ap:baselines}

In this section, we list all the baselines we compared our proposal with. The selected baselines are:

\noindent\textbf{Naive}: it is used as the lower bound, and it trains the models on the tasks without any strategy to fight the CF. 

\noindent\textbf{Cumulative}: this approach cumulatively trains the models using a dataset built up by merging all the tasks up to the current one. It represents the upper bound since all the training samples are saved in the external memory.

\noindent\textbf{Experience Replay} \cite{chaudhry2019tiny, buzzega2021rethinking}: it is a simple rehearsal approach, in which samples from the memory are used along with the ones from the current dataset to train the model. It has a fixed-sized memory populated with samples from the current task once it is over, discarding past samples to make room for newer ones.

\noindent\textbf{Greedy Sampler and Dumb learner} (GDumb) \cite{prabhu2020gdumb}: it was proposed for questioning the advantages of CL, and it simply avoids training the model when a new task is collected, which is just used to fill up the memory. The just-filled memory is then used to train a new model from scratch when needed.  

\noindent\textbf{Experience Replay with Asymmetric Cross-Entropy} (ER-ACE) \cite{caccia2022new}: initially proposed for contrasting CF in an Online CL scenario, it was also adapted for CIL. It uses a disjointed cross-entropy loss to leverage class unbalancing. 
    
\noindent\textbf{Separated Softmax for Incremental Learning} (SS-IL) \cite{ahn2021ss}: it mitigates the CF using Knowledge Distillation on a task-wise basis, in addition to a modified cross-entropy loss to learn patterns from the current task. 

\noindent\textbf{Gradient Episodic Memory} (GEM) \cite{lopez2017gradient}: it uses the external memory to calculate gradients associated with past tasks and uses them to move the current one in a region of the space that satisfies both the current task as well as past ones. It does so by minimizing a quadratic problem, whose computational complexity scales exponentially with the number of samples in the memory.

\noindent\textbf{Dark experience replay} (DER) \cite{buzzega2020dark}: the model is regularised by, at the same time, augmenting the current batch using past samples and regularizing the logits using the MSE distance loss between the logits obtained using the current model and the ones from the past model. The approach implements a classifier with a fixed number of classes, trained all at the same time.

\noindent\textbf{Regular Polytope Classifier} (RPC) \cite{pernici2021class}: the idea is to fight the CF by using a fixed number of equidistant and not learnable classifiers, and to learn only the backbone. To avoid setting the number of classes in advance we add a projection layer, which produces a vector containing 1000 features, between the backbone and the classifier layer, fixing the number of learnable classes to 1001.

\noindent\textbf{Loss Decoupled} (ER-LODE) \cite{liang2023loss}: it decouples the classification loss creating two components conditional on whether the samples are from the current task or not. It uses an external memory and combines the losses by weighting them using a scaling factor which depends on the number of seen classes.  

\noindent\textbf{Logits Distillation} (LD): we also propose a simple baseline called Logits Distillation, in which an external class-balanced memory contains images used to regularize the logits with a KL distance (weighted by a constant $\alpha$). To do so, a copy of the model, used as the teacher model, is created at the beginning of the training. Moreover, we regularize future logits as in our proposed method (Section \ref{sec:mbm}). This baseline is useful for studying the effect of future regularisation without involving any classification loss over past samples.

\section{Hyperparameters selection}
\label{ap:hps}
\begin{table}[h]
\centering
\resizebox{\linewidth}{!}{%
\begin{tabular}{|l|l|cccc|}
\hline
\multirow{2}{*}{Dataset} & \multirow{2}{*}{Method} & \multicolumn{4}{c|}{Memory size} \\ \cline{3-6} 
 &  & \multicolumn{1}{c|}{200} & \multicolumn{1}{c|}{500} & \multicolumn{1}{c|}{1000} & \multicolumn{1}{c|}{2000} \\ \hline
\multirow{4}{*}{C10-5} & DER & \multicolumn{1}{c|}{$\alpha=0.1, \beta=0.5$} & \multicolumn{1}{c|}{$\alpha=0.5, \beta=0.8$} & \multicolumn{1}{c|}{$\alpha=0.5, \beta=0.1$} &  \multicolumn{1}{c|}{$\alpha=0.2, \beta=0.8$} \\ \cline{2-6} 
& LODE & \multicolumn{1}{c|}{$\rho=0.2$} & \multicolumn{3}{c|}{$\rho=0.5$} \\ \cline{2-6} 
 & LD & \multicolumn{4}{c|}{$\alpha = 0.1$} \\ \cline{2-6}
 & \nrega & \multicolumn{1}{c|}{$\lambda=0.1$} & \multicolumn{2}{c|}{$\lambda=0.25$} &  \multicolumn{1}{c|}{$\lambda=0.5$} \\ \cline{1-6} \cline{1-6}
\multirow{4}{*}{C100-10} & DER & \multicolumn{1}{c|}{$\alpha=0.1, \beta=0.8$} & \multicolumn{1}{c|}{$\alpha=0.1, \beta=1$} & \multicolumn{2}{c|}{$\alpha=0.1, \beta=0.8$} \\ \cline{2-6}
& LODE & \multicolumn{1}{c|}{$\rho = 0.1$} & \multicolumn{2}{c|}{$\rho = 0.2$} & \multicolumn{1}{c|}{$\rho = 0.5$}\\ \cline{2-6}
& LD & \multicolumn{2}{c|}{$\alpha = 1$} & \multicolumn{2}{c|}{$\alpha = 0.75$} \\ \cline{2-6}
& \nrega & \multicolumn{1}{c|}{$\lambda=0.1$} & \multicolumn{1}{c|}{$\lambda=0.5$} & \multicolumn{1}{c|}{$\lambda=0.25$} &  \multicolumn{1}{c|}{$\lambda=1$} \\ \cline{1-6}
%
\end{tabular}%
}
\caption{The best combination of hyper-parameters for each approach when training on CIFAR-like scenarios.}
\label{table:hyperparameters}
\end{table}
\begin{table}[h]
\centering
\resizebox{0.6\linewidth}{!}{%
\begin{tabular}{|l|l|cc|}
\hline
\multirow{2}{*}{Dataset} & \multirow{2}{*}{Method} & \multicolumn{2}{c|}{Memory size} \\ \cline{3-4} 
 &  & \multicolumn{1}{c|}{2000} & \multicolumn{1}{c|}{5000} \\ \hline
\multirow{3}{*}{TyN-10} & DER & \multicolumn{2}{c|}{$\alpha=0.1, \beta=0.2$} \\ \cline{2-4} 
& LODE & \multicolumn{1}{c|}{$\rho=0.1$} & \multicolumn{1}{c|}{$\rho=0.5$} \\ \cline{2-4} 

& \nrega & \multicolumn{1}{c|}{$\lambda=0.1$} & \multicolumn{1}{c|}{$\lambda=0.05$}  \\ \hline
\multirow{3}{*}{TyN-20} & DER & \multicolumn{1}{c|}{$\alpha=0.1, \beta=0.5$} & \multicolumn{1}{c|}{$\alpha=0.1, \beta=0.1$} \\ \cline{2-4} 
& LODE & \multicolumn{1}{c|}{$\rho = 0.1$} & \multicolumn{1}{c|}{$\rho = 0.2$} \\ \cline{2-4} 

& \nrega & \multicolumn{1}{c|}{$\lambda=0.5$} & \multicolumn{1}{c|}{$\lambda=0.1$}  \\ \hline
\multirow{3}{*}{TyN-50} & DER & \multicolumn{1}{c|}{$\alpha=0.1, \beta=0.2$} & \multicolumn{1}{c|}{$\alpha=0.1, \beta=0.1$} \\ \cline{2-4} 
& LODE & \multicolumn{2}{c|}{$\rho=0.2$} \\ \cline{2-4} 
& \nrega & \multicolumn{1}{c|}{$\lambda=0.05$} & \multicolumn{1}{c|}{$\lambda=0.1$}  \\ \hline
\end{tabular}%
}
\caption{The best combination of hyper-parameters for each approach when training on Tiny ImageNet scenarios.}
\label{table:hyperparameters_tim}
\end{table}
To select the best hyperparameters, we fixed the number of epochs, the batch size, and the optimizer parameters. Then, we trained a model for each combination of hyperparameters, and we selected the best set by evaluating the results on a development split containing $10\%$ of the training samples. For each method, the hyperparameters are: 

\begin{itemize}
    \item DER \cite{buzzega2020dark}: this approach has two hyperparameters. The first controls the classification loss of past samples, $\alpha: [0.1, 0.2, 0.5, 0.8, 1.0]$, while the second controls the logits distillation loss $\beta: [0.1, 0.2, 0.5, 0.8, 1.0]$. 
    \item LODE \cite{liang2023loss}: the only parameter $\rho$ controls the weight of the proportion between the number of new classes over the number of past ones. The resulting value is used to weigh one of the losses. Based on the findings shown in the paper, the search space is $\rho: [0.1, 0.2, 0.5, 0.8, 1]$.
    \item LD: it has just one parameter, which controls the strength of the logit regularisation. The search space is $\alpha: [0.1, 0.25, 0.5, 0.75, 1]$
    \item \nrega: our proposal has only one hyperparameter, which controls the strength of margin regularisation, and the search space is $\lambda: [1, 0.5, 0.25, 0.1, 0.05, 0.025, 0.01]$.
    
\end{itemize}

\noindent The best parameters, for each combination of scenario and memory, are shown in Table \ref{table:hyperparameters} and Table \ref{table:hyperparameters_tim}.

\section{Metrics}
\label{ap:metrics}

To evaluate the efficiency of a CL method, we use two different metrics proposed in \cite{diaz2018don}. The first one, called Accuracy, shows the final accuracy obtained across all the tasks' test splits, while the second one, called Backward Transfer (BWT), measures how much of that past accuracy is lost during the training on upcoming tasks. Both metrics are important, and a trade-off must be achieved to balance plasticity (high accuracy on current task) and stability (low forgetting). However, even if both are important to compare CL approaches, the accuracy usually has a bigger weight in the overall evaluation. 
To calculate the metrics, we use a matrix $\mathbf{R} \in \mathbb{R}^{\text{N} \times \text{N}}$, in which an entry $\mathbf{R}_{i, j}$ is the test accuracy obtained on the test split of the task $j$ when the training on the task $i$ is over. Using the matrix $\mathbf{R}$ we calculate the metrics as:
\begin{multicols}{2}
\begin{equation*}
 \text{ACC} = \frac{1}{\text{\text{N}}}\sum_{j=1}^\text{\text{N}} \mathbf{R}_{\text{\text{N}}, j}
\end{equation*}\break
\begin{equation*}
 \text{BWT} = \frac{\sum_{i=2}^{\text{\text{N}}} \sum_{j=1}^{i-1} (\mathbf{R}_{i, j} - \mathbf{R}_{j, j})}{\frac{1}{2} \text{\text{N}} (\text{\text{N}}-1)} \,.
\end{equation*}
\end{multicols}
%
 Both metrics are important to evaluate a CL method since a low BWT does not imply that the model performs well, especially if we also have a low Accuracy score because, in that case, it means that the approach regularizes too much during the training, leaving no space for learning new tasks. In the end, the combination of the metrics is what we need to evaluate.

\section{Memory overhead}
\label{ap:overhead}
To calculate the memory overhead, we need to count additional parameters each approach requires. As additional parameters, we consider the ones that must be stored in addition to the base model and the tasks' head.

For a generic rehearsal approach, the number of additional floats to store is given by $O = W \times H \times C \times \lvert \mathcal{M} \rvert$, in which $(C, H, W)$ are the sizes of the images in the dataset and $\mathcal{M}$.  

For our proposal, in addition to the parameters $O$, we also count the additional parameters in the classifier head \nheada. Having the input of the head size equal to I and the current task $T$, we have: 

\begin{equation}
    O_{\nheada}^T = O + \text{I} \times \sum_{i=1}^{T} (i-1) \lvert \mathrm{Y}^i \rvert  
\end{equation}

where $\lvert \mathrm{Y}^i \rvert$ is the number of classes in task i. For example, when measuring $O_{\nheada}^T$ at the end of the final C10-5 task, we have that $O_{\nheada}^T = O + \text{I} \times 2 \times 15$, since $ \lvert \mathrm{Y}^i \rvert  = 2$ for each task. For ResNet20, we have $I = 64$, which results in  $O_{\nheada}^T = O + 64 \times 2 \times 15 = O + 1920$. So, in this case (10 tasks), all the additional parameters contained in the scaler heads are less than the ones used to store a single image in the memory, making the overhead of \nheada negligible. 

\section{Hardware settings and computational time}
We run all the experiments on two different machines. The first is a laptop with an Nvidia Geforce RTX 4080, while the second is a shared cluster having multiple A100, from which we used one. 

The time needed to run the experiments varies based on the number of tasks and the dimension of the dataset. The approximate time needed to run the experiments also varies from one method to another. However, the best methods require, approximately, the following computational time per experiment: 

\begin{itemize}
    \item C10-5: 30m
    \item C100-10: 45m
    \item TyM-10: 1.5h
    \item TyM-20: 3h
    \item TyM-50: 5h
\end{itemize}

Such methods are ER-LODE, ER-ACE, REPLAY, LD, and DER. Regarding our approach, it requires approximately 1.2 times the reported complexity. GEM and SS-IL require a time which grows with the number of tasks, making them almost infeasible for harder scenarios. 
    
\section{Additional experiments}
\label{app:additional_experiments}
\subsection{\nheada time overhead and additional parameters growing law}
\label{app:time}

\begin{figure}[h]
\centering
\begin{minipage}[t]{.45\columnwidth}
  \centering
  \includegraphics[width=1\columnwidth]{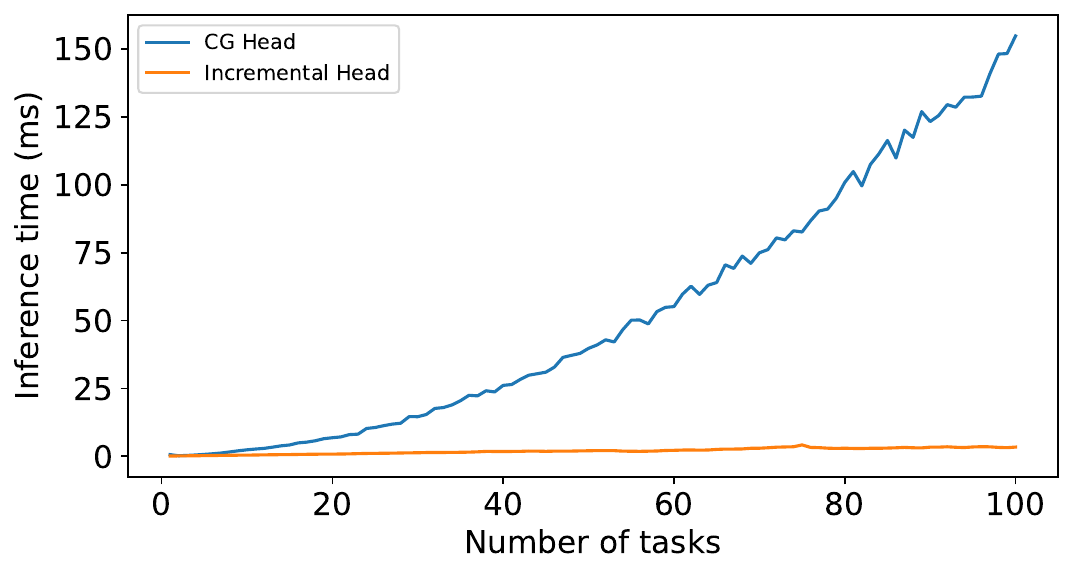}
  \captionof{figure}{The empirically estimated inference time for our proposal \nheada and the standard Incremental Head with an increasing number of tasks.}
  \label{fig:time}
\end{minipage}%
\hfill
\begin{minipage}[t]{.45\textwidth}
  \centering
  \includegraphics[width=1\columnwidth]{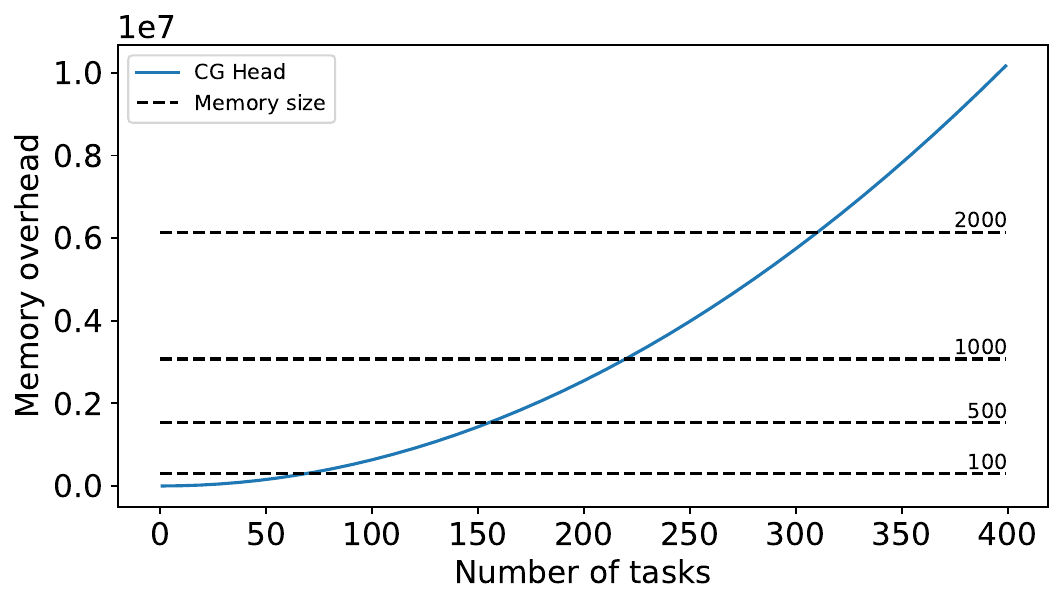}
  \captionof{figure}{Number of additional parameters added by \nheada compared to the memory overhead. Results obtained for a CIFAR-like scenario, when using a ResNet and 2 classes per task.}
  \label{fig:growing}
\end{minipage}
\end{figure}

In this section, we analyse the time overhead of the proposed head, \nheada, as well as the growing law of the number of additional parameters it introduces.

To calculate the time, we generated 20 virtual tasks, meaning that no real images are used, but the growing number of tasks is simulated to adapt the models. To measure the time, in milliseconds, we used the Event functionality in the Pytorch framework \footnote{\href{https://pytorch.org/docs/stable/generated/torch.cuda.Event.html}{https://pytorch.org/docs/stable/generated/torch.cuda.Event.html}}. To have better statistics, we simulate a forward pass in the whole model, and multiple forward (100) passes into the head we are testing. This process is repeated for each task and the average elapsed time is saved. Figure \ref{fig:time} shows that, for our approach, such time indeed grows, but this growth is linear with the number of tasks and not quadratic as the number of heads. Moreover, being the time in milliseconds, we can say that the time overhead introduced by our proposal is contained.

Regarding the growing law of the additional parameters added by our proposal \nheada, we observe that, for a generic task index $t > 1$, we need to add a head for each past task, which are $t-1$. This must be summed to the number of heads already added, leading us to the recursive formula $G(t) = t-1 + G(t-1)$, which gives us the number of additional heads for a task $t$. Hence, the number of heads can be calculated as: 
\begin{equation}
    G(t) = \frac{(1 + (t -1)) (t -1)}{2}
\end{equation}
which grows quadratically. Supposing a CIFAR-like dataset, which has images of size $3 \times 32 \times 32$, 2 classes per each task, and a ResNet-like architecture, Figure \ref{fig:growing} shows us the parameter growing law compared to the additional parameters added by the external memory. It shows that our approach indeed adds a quadratic number of parameters but this value surpasses the dimension of the memory only when the number of tasks is very high. Recalling that more tasks require larger memory to effectively fight the CF, we conclude that the \nheada 's parameters are negligible when compared to the size of the memory, even when the dimension of the images is contained. 

\subsection{On scaling the Sigmoid function}
\begin{figure}[h]
\centering
\scalebox{0.85}{
    \subfloat{
    \includegraphics[width=0.5\linewidth]{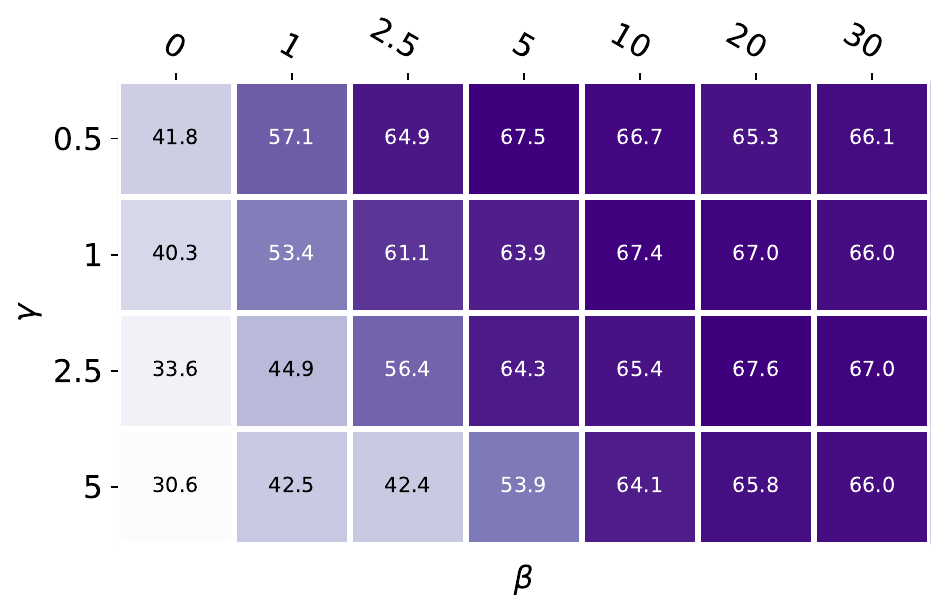}
    \label{fig:simgoid_1}
    }
    \hfill
    \subfloat{
    \includegraphics[width=0.5\linewidth]{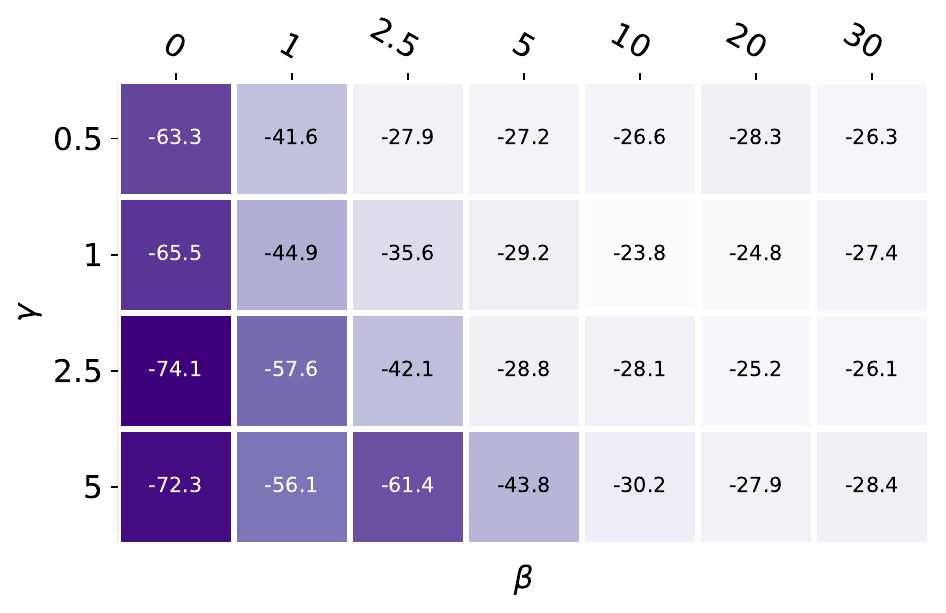}
    \label{fig:simgoid_2}
    }
    }
    \caption{How the accuracy (left) and the BTW (right) scores are affected when varying the parameters of the scaler function proposed in Section \ref{sec:scaler}. The results were obtained on ResNet-20 trained on C10-5 with a memory size set to 500.}
    \label{fig:simgoid}
\end{figure}

In this section, we analyse how the scale ($\gamma$) and the offset ($\beta$) values used in the scaling function, proposed in Section \ref{sec:scaler}, affect the results. Intuitively, a strong offset $\beta$ is necessary, since a high value will preserve past logits by not gating them at the beginning of the training, creating a model which is capable of outputting a distribution that resembles the one obtained by the model before starting the training on a new task. As shown in Fig. \ref{fig:simgoid}, for the C10-5 scenario using a memory size of 500, a low offset value leads to a lower accuracy, which is partially recovered when the scaling value is lower than $1$.  Overall, all the results stop improving when the offset value is higher or equal to $20$.  
This is expected since a low offset does not guarantee that the model is regularised using the correct output distributions. 
Regarding the scaling value, which controls the smoothness of the Sigmoid curve, it behaves like a balance factor when the offset is below or equal to $2.5$, but it becomes less impacting as the offset gets higher, having a negligible impact when the latter reaches $20$. For this reason, in our experiment, we fixed the offset to 10 and the scale to 1. 
\begin{figure}
\centering
\begin{minipage}[t]{.4\columnwidth}
  \centering
  \includegraphics[width=1\columnwidth]{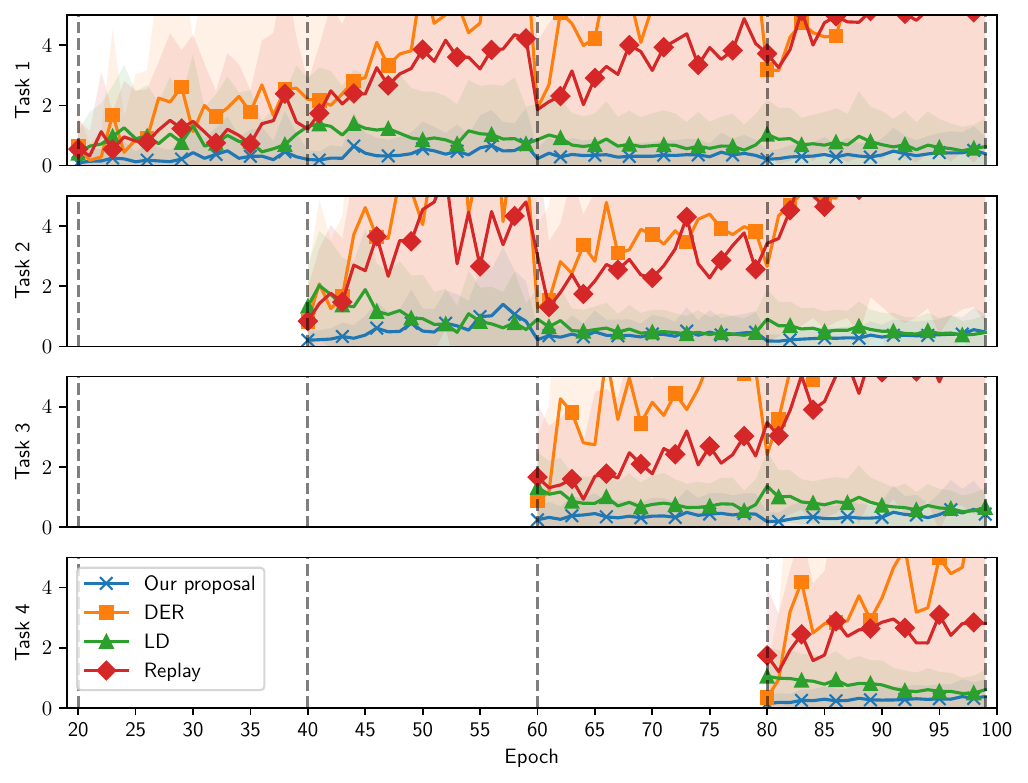}
  \captionof{figure}{The images show the KL distance between the distributions predicted for past samples when a new task is collected and while training on it. The results are calculated over the whole test set of the C10-5 scenario, using the best hyperparameters and a memory size of 500.}
  \label{fig:future}
\end{minipage}%
\hfill
\begin{minipage}[t]{.49\textwidth}
  \centering
  \includegraphics[width=0.9\columnwidth]{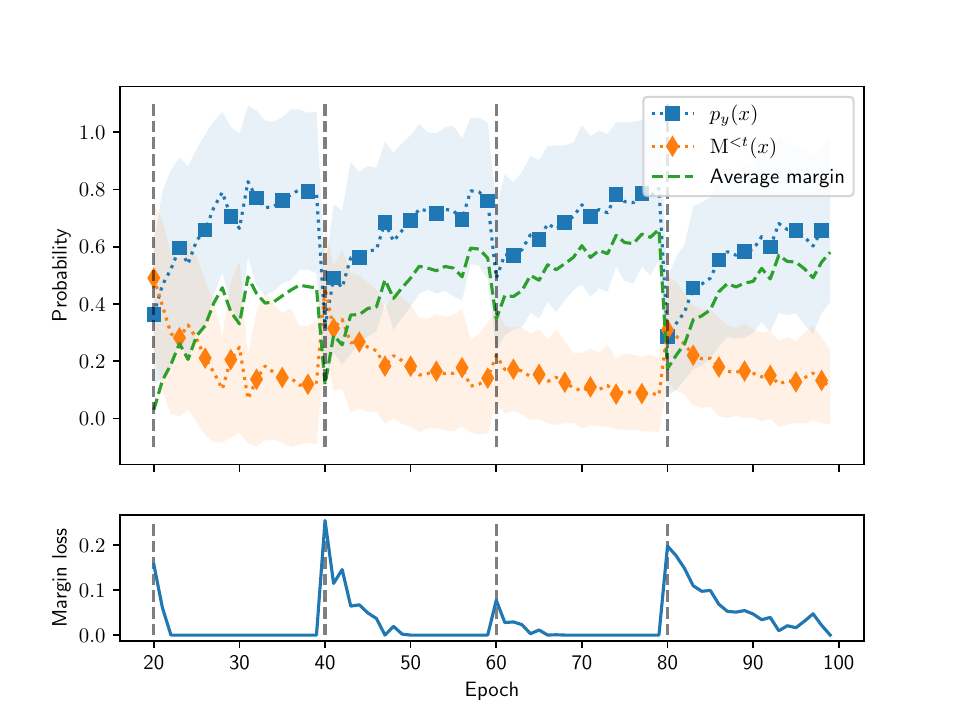}
  \captionof{figure}{The figure (up) shows how the past logits, as well as the ground truth ones and the margin loss (down), vary during the training from the second task onward. Results were obtained on ResNet-20 trained on C10-5, with 500 samples saved in the memory and our proposal as the regularisation method.}
  \label{fig:margin_loss}
\end{minipage}
\end{figure}

\subsection{Regularising future classes} 

Using past samples as training ones seems important to avoid learning current classes $\mathrm{Y}^t$ using only current task samples, called positive samples, which inevitably leads to CF. 
Moreover, we know that using such samples to augment the training procedure could lead to overfitting, increasing forgetting. Instead, we advocate that training on such samples is unnecessary and that future classes, concerning a past class $t'<t$, can be easily regularised instead of trained, 
that improves both the plasticity and the stability. Here, we show this aspect by evaluating how the predictive ability of the model changes while training on newer tasks. Each time a new task $t$ is collected, we calculate the distribution $p^t_0(x) \in \mathbb{R}^{\lvert \mathcal{Y}^{t} \rvert}$, for all $x \in \mathcal{D}^{t'}_{\text{test}}$  and for each $t' < t$. Then, we evaluate how such distributions change while training on $t$ using the KL divergence between the current distributions and $p^t_0(x)$. 

Figure \ref{fig:future} shows such results, obtained on C10-5 trained using a memory size of 500, for DER, Replay, LD, and our proposal. DER, which uses a fixed number of output classes, regularised and trained simultaneously, diverges. Such behaviour is unsurprising as the approach trains all the future classes using only positive samples and then tries to keep the output of the model fixed while training it, negatively impacting the stability. Consequently, it must strike a delicate and hard-to-tune balance to work efficiently. Even then, ER-ACE and ER-LODE achieve better results. Since we observe the same divergence when using Replay, we conclude that DER regularisation has a negligible effect on future classes. On the other hand, LD can preserve the model's predictive ability by keeping the divergence low during training. However, as seen in the main results, it lacks plasticity, resulting in low scores. Instead, our main proposal removes the overfitting by not training on past samples and by using a regularisation schema that allows for more plasticity without detriment to the stability. Ultimately, our main proposal is the only rehearsal-regularisation approach capable of preserving the model's predictive capability while allowing for a higher degree of plasticity. These aspects combined lead to higher results overall, as already shown in the main results.

\subsubsection{\nrega regularisation effects}
Figure \ref{fig:margin_loss} shows how the probabilities change when training the model using the proposed \nreg regularisation approach. As we can see in the top image, the ground truth probabilities constantly increase while, at the same time, the maximum probabilities from past tasks decrease. The regularisation loss reaches zero after a few epochs, showing the regularisation ability of the proposed approach. Despite that, the maximum past value associated with current training samples keeps reducing even after the loss reaches zero (due to the soft constraint imposed by the Equation \eqref{eq:margin_loss}). Such decreasing does not affect the output produced for past samples since the constraint imposed by the regularisation term is already respected, zeroing the regularisation term. 

\end{document}